\documentclass[conference]{IEEEtran}
\IEEEoverridecommandlockouts
\usepackage{cite}
\usepackage{amsmath,amssymb,amsfonts}
\usepackage{graphicx}
\usepackage{textcomp}
\usepackage{xcolor}
\usepackage{booktabs}
\usepackage{amsmath}
\usepackage{fontawesome5}
\usepackage{placeins}
\usepackage{graphicx}
\usepackage{float}
\usepackage{tcolorbox}
\usepackage{amsmath}    
\usepackage{algorithm}
\usepackage{algpseudocode}
\usepackage{hyperref}

\def\BibTeX{{\rm B\kern-.05em{\sc i\kern-.025em b}\kern-.08em
    T\kern-.1667em\lower.7ex\hbox{E}\kern-.125emX}}

\IEEEaftertitletext{\vspace{-2\baselineskip}\begin{center}\textit{Published in: 2025 IEEE International Conference on Big Data (BigData).
DOI: \href{https://doi.org/10.1109/BigData66926.2025.11401919}{10.1109/BigData66926.2025.11401919}}\end{center}}

\begin{document}

\title{Rethinking Ground Truth: A Case Study on Human Label Variation in MLLM Benchmarking\\

\thanks{This research is funded by the Bavarian Research Institute for Digital Transformation (bidt), and was conducted within the ToxicAInment research project.}
\thanks{\copyright~2025 IEEE. Personal use of this material is permitted. Permission from IEEE must be obtained for all other uses, in any current or future media, including reprinting/republishing this material for advertising or promotional purposes, creating new collective works, for resale or redistribution to servers or lists, or reuse of any copyrighted component of this work in other works.}
}
\author{\IEEEauthorblockN{1\textsuperscript{st} Tomas Ruiz}
\IEEEauthorblockA{\textit{Computational Social Sciences} \\
\textit{LMU Munich}\\
Munich, Germany \\
t.ruiz@lmu.de}
\and
\IEEEauthorblockN{2\textsuperscript{nd} Tanalp Agustoslu}
\IEEEauthorblockA{\textit{Computational Linguistics} \\
\textit{LMU Munich}\\
Munich, Germany \\
t.agustoslu@campus.lmu.de}
\and
\IEEEauthorblockN{3\textsuperscript{rd} Carsten Schwemmer}
\IEEEauthorblockA{\textit{Computational Social Sciences} \\
\textit{LMU Munich}\\
Munich, Germany \\
carsten.schwemmer@lmu.de}
}

\maketitle

\begin{abstract}
Human Label Variation (HLV), i.e. systematic differences among annotators’ judgments, remains underexplored in benchmarks despite rapid progress in large language model (LLM) development. 
We address this gap by introducing an evaluation protocol for multimodal large language model (MLLM) benchmarking that explicitly accounts for two conditions: (1) human label agreement and (2) disagreement.
We apply this protocol to two state-of-the-art MLLM families (Gemma 3, Qwen 2.5 VL) using non-aggregated human annotations from a social media content classification dataset.
Across tasks, we find that larger models tend to perform best on high-agreement subsets, yet often underperform medium-sized models when human disagreement is high, indicating that parameter count alone does not determine sensitivity to ambiguity and subjectivity. 
These results show that benchmarks based solely on consensus labels can overstate model capabilities in such domains and that incorporating human label variation yields more realistic and robust assessments of MLLMs in content moderation pipelines.
\end{abstract}


\section{Introduction}
\noindent Social media platforms host large volumes of political and entertainment content, including toxic material that can harm psychological well-being and degrade public discourse~\cite{b6,b9,b11}. Moderating such content at scale is challenging because it spans multiple modalities (text, images, audio, and video) whose interplay can obscure a post’s true intent~\cite{b17}. Multimodal large language models (MLLMs) provide a unified latent space for diverse inputs and promise more accurate and effective content understanding and moderation~\cite{b18, b19, b20}.

Most existing benchmarks \cite{b21, b22} have demonstrated that MLLMs can solve complex tasks such as mathematical reasoning, object detection, and multimodal understanding, often achieving human-like performance. These evaluations are typically conducted by reducing tasks to binary or multiclass classification formats, which assume the existence of objective ground-truth labels and rely on human-annotated corpora, or annotations produced by LLMs or MLLMs for verification. Some benchmarks additionally include expert review to correct obvious annotation errors~\cite{b14, b23}. 

However, many content moderation tasks involve substantial ambiguity and subjectivity, where annotators legitimately disagree. In such settings, majority voting can obscure Human Label Variation (HLV) and produce a single ``ground truth'' that does not reflect the spectrum of human judgments. For example, in the multimodal benchmark MMHS150K \cite{b24}, models that included a vision modality performed worse than text-only models, in part due to noisy labels in a subjective task. Rather than treating disagreement as noise to be filtered out, recent work on HLV argues for preserving this variation and evaluating models against it~\cite{b2, b5, b8, b31}. 

In this paper, we follow this perspectivist view and use disaggregated labels during evaluation to assess how well MLLMs align with human judgments overall. Our central question is: \textit{\textbf{How can an evaluation protocol for MLLM benchmarking explicitly account for human label variation in both agreement and disagreement settings to more accurately reflect model performance?}} 

Our contributions are threefold:
\begin{itemize}
    \item We propose an evaluation protocol that uses inter-annotator agreement (Krippendorff’s Alpha) to separate evaluation into two distinct subsets: an agreement subset and a disagreement subset.
    \item Using non-aggregated annotations, we benchmark two state-of-the-art MLLM families to show how the protocol reveals nuanced patterns in model performance and sensitivity to human disagreement.
    \item We make our benchmarking code publicly available to facilitate further research\footnote{\url{https://github.com/agustoslu/simple-inference-benchmark}}.
\end{itemize}

At a high level, our evaluation protocol partitions questions into agreement and disagreement subsets based on Krippendorff’s Alpha and applies classification metrics to the former and distribution-sensitive probabilistic metrics to the latter. We instantiate this protocol on the PoliTok-DE content moderation dataset using the Gemma 3 and Qwen 2.5 VL model families (Section~\ref{sec:methods}).

\begin{table*}[h!]
  \caption{
  \textbf{Agreement Subset Evaluation}\\Higher is better. Bold: best value by model family.}
  \centering
  \small
  \setlength{\tabcolsep}{4pt}
  
  \begin{tabular}{lcccccc}
  \toprule
  \textbf{Question} & \multicolumn{3}{c}{\textbf{is\_political}} & 
    \multicolumn{3}{c}{\textbf{is\_saxony}} \\
  \cmidrule(lr){2-4} \cmidrule(lr){5-7}
  \textbf{Model} & \textbf{Precision \(\uparrow\)} & \textbf{Recall \(\uparrow\)} & \textbf{F1 \(\uparrow\)} &
    \textbf{Precision \(\uparrow\)} & \textbf{Recall \(\uparrow\)} & \textbf{F1 \(\uparrow\)} \\
  \midrule
  \raggedright Gemma-3 4B It & 0.84 & 1.00 & 0.91 & 0.27 & 0.93 & 0.42 \\
  \raggedright Gemma-3 12B It & 0.93 & 0.97 & \textbf{0.95} & 0.71 & 0.81 & \textbf{0.76} \\
  \raggedright Gemma-3 27B It & 0.92 & 0.98 & \textbf{0.95} & 0.66 & 0.84 & 0.74 \\
  \midrule
  \raggedright Qwen 2.5 VL 3B Instruct & 0.91 & 0.97 & 0.94 & 0.36 & 0.91 & 0.52 \\
  \raggedright Qwen 2.5 VL 7B Instruct & 0.99 & 0.72 & 0.83 & 0.31 & 0.56 & 0.40 \\
  \raggedright Qwen 2.5 VL 72B Instruct & 0.96 & 0.95 & \textbf{0.95} & 0.82 & 0.86 & \textbf{0.84} \\
  \bottomrule
  \end{tabular}
  \label{tab:agreement_metrics}
\end{table*}

\begin{table*}[h!]
  \caption{
    \textbf{Disagreement Subset evaluation}\\Lower is better. Bold: best value by model family.}
  \centering
  \small
  \setlength{\tabcolsep}{4pt}
  
  \begin{tabular}{lcccccc}
  \toprule
  \textbf{Question} & \multicolumn{2}{c}{\textbf{is\_intolerant}} & 
    \multicolumn{2}{c}{\textbf{is\_hedonic}} & 
    \multicolumn{2}{c}{\textbf{is\_eudaimonic}} \\
  \cmidrule(lr){2-3} \cmidrule(lr){4-5} \cmidrule(lr){6-7}
  \textbf{Model} & \textbf{Brier \(\downarrow\)} & \textbf{JSD \(\downarrow\)} &
    \textbf{Brier \(\downarrow\)} & \textbf{JSD \(\downarrow\)} &
    \textbf{Brier \(\downarrow\)} & \textbf{JSD \(\downarrow\)} \\
  \midrule
  \raggedright Gemma-3 4B It & 0.37 & 0.26 & 0.38 & 0.27 & 0.27 & 0.20 \\
  \raggedright Gemma-3 12B It & \textbf{0.17} & \textbf{0.13} & \textbf{0.29} & \textbf{0.21} & 0.26 & \textbf{0.19} \\
  \raggedright Gemma-3 27B It & 0.31 & 0.22 & 0.37 & 0.26 & \textbf{0.25} & \textbf{0.19} \\
  \midrule
  \raggedright Qwen 2.5 VL 3B Instruct & 0.23 & 0.17 & 0.29 & 0.22 & 0.25 & 0.19 \\
  \raggedright Qwen 2.5 VL 7B Instruct & \textbf{0.12} & \textbf{0.10} & 0.28 & 0.21 & 0.25 & 0.19 \\
  \raggedright Qwen 2.5 VL 72B Instruct & 0.20 & 0.15 & \textbf{0.26} & \textbf{0.19} & \textbf{0.23} & \textbf{0.17} \\
  \bottomrule
  \end{tabular}
  \label{tab:disagreement_metrics}
\end{table*}

\section{Related Work}
Before introducing our evaluation protocol, we situate it within prior work on multimodal content moderation benchmarks and on Human Label Variation in subjective NLP tasks.

\subsection{Benchmark Datasets on Content Moderation}
\noindent The detection and reduction of harmful content has been an active area of research for many years. Early work primarily focused on single-modality evaluations, analyzing either text or visual elements in isolation \cite{b25}. With the rise of social media and the prevalence of memes, live streams, and video content, this challenge has evolved toward genuinely multimodal settings. To spur progress, several benchmark datasets have been introduced: Fakeddit \cite{b26} for multimodal fake news detection, MMHS150K \cite{b24} for hate speech in tweets with images, and the Hateful Memes Challenge \cite{b27} for testing multimodal reasoning where the combination of text and image creates hateful content. In most of these benchmarks, labels are aggregated into a single ground truth and models are evaluated on standard classification metrics. Other work has proposed ensembles and hybrid architectures to cope with multimodal harmful content \cite{b24, b28}. For video understanding in political contexts, CLARITY targets political question evasions \cite{b29}, while PoliTok-DE focuses on intolerant content in election-related videos on social media~\cite{b30}.

\subsection{Human Label Variation}
\noindent The concept of Human Label Variation (HLV) has caused a paradigm shift in NLP, giving rise to studies on inter-annotator agreement, the benefits of non-aggregation, and what has been called the ``perspectivist manifesto''~\cite{b5, b8, b32}. HLV challenges the assumption that there is a single, objective ground truth in subjective domains such as hate speech detection. Aggregating collected labels can introduce majority biases and obscure the real-world complexity of the task. As a result, models trained and evaluated on aggregated labels may fail to reflect inherent human uncertainty and instead reproduce a majority perspective.
These tasks often involve multiple forms of ambiguity, including linguistic (e.g., pronoun reference or sarcasm), sociocultural (phrases acceptable in some groups but offensive in others), and modality-specific ambiguities in multimodal content such as memes. Rather than discarding low-agreement labels, HLV approaches employ either soft labels or disaggregated labels to capture a broader spectrum of human judgments. Reliability metrics such as Krippendorff’s Alpha and Fleiss’ Kappa are used to quantify inter-annotator agreement~\cite{b33}, while probabilistic metrics (e.g., KL or Wasserstein distances) evaluate how well models capture this spectrum of perspectives~\cite{b37}.
Together, these findings directly motivate our evaluation protocol, which assesses models on their capacity to deal with annotation disagreements by explicitly separating agreement and disagreement subsets and using metrics that preserve label distributions.

\section{Methods}
\label{sec:methods}

\subsection{Dataset}
\noindent For our case study, we use the PoliTok-DE dataset, which contains multimodal TikTok posts related to the 2024 Saxony state election in Germany~\cite{b30}. To our knowledge, it is the only available video dataset with non-aggregated human labels for intolerant content, making it ideal for simulating a real-world content moderation setting. The dataset contains 195{,}373 posts in total; of these, 18{,}842 were removed by the platform. Humans annotated 300 of the removed posts on five different questions, yielding 1{,}500 labeled examples (300 posts $\times$ 5 questions), which we use for benchmarking.
The five annotation questions used in this study are: \textbf{is\_political}, indicating whether the content references political topics, actors, ideologies, or events; \textbf{is\_saxony}, specifying whether the content relates directly to the Saxony 2024 state elections or regional political issues; \textbf{is\_intolerant}, flagging content that expresses or implies discriminatory, hateful, or exclusionary attitudes toward individuals or groups; \textbf{is\_hedonic\_entertainment}, identifying content primarily intended to entertain, amuse, or emotionally engage through sensationalism, parody, or provocation; and \textbf{is\_eudaimonic\_entertainment}, marking content that encourages thoughtful reflection, raises awareness, or promotes civic or moral engagement.
Some questions are more subjective and/or ambiguous than others, resulting in different levels of annotator disagreement.

\subsection{Evaluation Protocol}

\noindent To assess annotation reliability and structure our evaluation, we partition the dataset based on the degree of human agreement, quantified using Krippendorff’s Alpha ($\alpha$).
Perfect agreement yields $\alpha = 1$, chance-level agreement yields $\alpha = 0$, and systematic disagreement results in $\alpha < 0$.
Krippendorff’s Alpha accommodates missing annotations in contrast to metrics such as Fleiss’ Kappa that assume all annotators label every item.

Following the reliability threshold of 0.667 suggested by Krippendorff~\cite{b2}, we define two subsets: an \textbf{agreement subset} consisting of questions with \textit{$\alpha \geq 0.667$}, indicating high inter-annotator reliability; and a \textbf{disagreement subset} consisting of questions with \textit{$\alpha < 0.667$}, reflecting inherent ambiguity and/or subjectivity (\autoref{tab:krippendorff}). This split allows us to compare model behavior on consensus-like tasks and on inherently controversial or ambiguous ones within a single benchmark.

\begin{table}[h!]
  \caption{
    Questions by Evaluation Subset\\
    Sorted by descending Krippendorff's Alpha ($\alpha$).
  }
  \centering
  \begin{tabular}{lcc}
  \toprule
  \textbf{Question} & \textbf{$\alpha$} & \textbf{Subset} \\
  \cmidrule(lr){1-3}
  is\_political & 0.81 & Agree \\
  is\_saxony & 0.74 & Agree \\
  \midrule
  is\_hedonic\_entertainment & 0.55 & Disagree \\
  is\_intolerant & 0.48 & Disagree \\
  is\_eudaimonic\_entertainment & 0.38 & Disagree \\
  \bottomrule
  \end{tabular}
  \label{tab:krippendorff}
\end{table}

For the \textbf{agreement subset}, we derive pseudo-ground-truth labels by applying majority voting to human annotations.
Model labels are also obtained using majority voting across five model runs. 
We then report \textit{precision}, \textit{recall}, and \textit{F1-score} comparing the model labels with the pseudo-ground truth.

For the \textbf{disagreement subset}, we focus on assessing the alignment between model predictions and the empirical distribution of human labels by using two metrics: (1) \textit{Brier score} and (2) \textit{Jensen-Shannon Divergence (JSD)}.
Both metrics preserve the full distribution of human judgments, making them suited for evaluating model sensitivity to disagreement.
The Brier score measures model calibration, functioning as a mean squared error for predicted probabilities. A lower Brier score indicates that the model’s predicted probabilities (e.g. 60\% intolerant) closely match the human label distribution (e.g. 3 out of 5 annotators labeled content as intolerant). It penalizes models that are confidently incorrect (e.g. predicting 100\% when annotators are evenly split). We compute it using soft labels by comparing the mean of the model’s predictions (averaged across five runs) with the mean of the human labels. The JSD measures the distance between the model’s predicted probability distribution and the empirical human label distribution. A lower JSD indicates that the model’s predictions better capture the shape of the human distribution.

\section{Results}
\paragraph{Agreement subset} We present the results for the agreement subset in \autoref{tab:agreement_metrics}, focusing on F1 score as our main summary metric.
Within the Qwen2.5-VL model family, the largest model (72B) performs best, as expected, while in the Gemma-3 family the largest model (27B) and medium-sized model (12B) achieve very similar performance.
Across both families, all models perform better on political content detection than on geographic detection (Question \textit{is\_saxony}), as seen by the substantially lower F1 scores for the latter task (e.g., 0.83 vs. 0.40 for the weakest models in each family).

\paragraph{Disagreement subset} The results for the disagreement subset are presented in \autoref{tab:disagreement_metrics}, where lower Brier and JSD values indicate better alignment with human label distributions.
Within the Qwen2.5-VL family, the largest model (72B) performs best in 4 out of 6 cases, while in the Gemma-3 family the medium-sized model (12B) matches or exceeds the largest model (27B) in 5 out of 6 cases by a large margin, indicating better calibration to the empirical distribution of human labels. 

At first glance, one might expect that larger models should consistently perform better across both subsets. Our results instead suggest that parameter count alone does not determine sensitivity to human disagreement. 
Taken together, these findings highlight that strong performance on consensus-based benchmarks may overstate the true capabilities of models in subjective domains. Incorporating human label variation into evaluation reveals the extent to which models capture the diversity of human judgment, offering a more realistic and robust framework for assessing multimodal model behavior in content moderation pipelines.

\section{Discussion}
\noindent The observed discrepancy between model performance on agreement versus disagreement subsets may stem from multiple factors. Here, we discuss two potential explanations and outline limitations and directions for future work.

\paragraph{Instruction Tuning and Alignment Tax} 
We used instruction-tuned models for consistent comparison across both families.
However, instruction tuning is known to increase model confidence (the so-called \textit{alignment tax})~\cite{b34, b35, b36}.
While this is beneficial for consensus tasks, it can be problematic when human uncertainty is high: higher confidence produces sharper logits and limits variance across runs.
This may explain why some models struggle to capture the distribution of human judgments on disagreement tasks, where lower confidence and more calibrated uncertainty would be appropriate.
Future work could directly compare base models (e.g., Gemma-3 PT) against their instruction-tuned counterparts to test whether the alignment tax contributes to reduced performance in the disagreement subset.

\paragraph{Shortcut Reasoning and Multimodal Understanding} 
Qualitative analysis of both human and AI explanations suggests that strong performance on high-agreement subsets may stem from reliance on explicit textual metadata (e.g., hashtags mentioned in the text) rather than full multimodal understanding, indicating that both humans and MLLMs often leverage surface cues rather than deep content reasoning. This type of shortcut further highlights why benchmarking performance on an agreement subset only can overstate model capabilities.

\paragraph{Limitations and Future Work}
Our study focuses on a single dataset (PoliTok-DE) and two contemporary MLLM families in a specific political content moderation context.
Future work could apply the same protocol to additional domains, languages, and model architectures to test the generality of our findings.

\section{Conclusion}
\noindent Accounting for Human Label Variation is essential for robust evaluation of MLLMs in subjective and ambiguous domains such as content moderation.
By partitioning evaluation into agreement and disagreement subsets based on inter-annotator reliability, our protocol shows that model size alone does not determine sensitivity to human disagreement: larger models excel on high-agreement tasks, whereas medium-sized models often better capture the distribution of human judgments.
These findings challenge the assumption that consensus-based benchmarks fully reflect model capabilities and highlight the need to consider disagreement-aware metrics when deploying MLLMs in real-world moderation pipelines.
Our evaluation protocol and public code provide a reusable foundation for more accurate and realistic benchmarking in these domains.

\appendices
\section{Experimental Details}

\subsection{Adaptive Frame Sampling}
\noindent We set a maximum frame budget $N_{\text{max}}$ per video to limit input size and prevent out-of-memory (OOM) errors. For a video with $T$ frames at $F$ fps, duration is $D = T / F$. If $D \leq N_{\text{max}}$, we sample at 1 FPS (every $F$-th frame); otherwise, we uniformly extract $N_{\text{max}}$ frames.
Formally, frame indices are defined as:

\[
I = \left\{ \left\lfloor k \cdot \text{step} \right\rfloor \;\middle|\; k = 0, 1, 2, \dots \right\},
\]
where
\[
\text{step} = 
\begin{cases} 
F & \text{if } D \leq N_{\text{max}}, \\[6pt]
T / N_{\text{max}} & \text{otherwise.}
\end{cases}
\]

This ensures that (i) short videos use 1 FPS, (ii) long videos are downsampled to $N_{\text{max}}$ frames, and (iii) frame counts are bounded across the benchmark\footnote{\url{https://github.com/tomasruizt/llm_app/blob/main/llmlib/llmlib/huggingface_inference.py\#L62}}.

\subsection{Sampling Parameters}
\noindent For both model families, we use the parameters presented in their \texttt{generation configurations} on HuggingFace\footnote{\url{https://huggingface.co/google/gemma-3-27b-it/tree/main}}\footnote{\url{https://huggingface.co/Qwen/Qwen2.5-VL-72B-Instruct/tree/main}}, while overriding the decoding parameters to \texttt{top-k=50}, \texttt{top-p=1.0}, and \texttt{temperature=0.7} across five stochastic runs. 

\subsection{Prompting}
\noindent We provide each model with the same annotation guidelines as humans, integrate metadata, and ask it to classify each video across all five questions in a single inference step using structured JSON output\footnote{\url{https://github.com/agustoslu/simple-inference-benchmark/blob/main/bench_lib/src/bench_lib/prompts/prompt.txt}}.

\subsection{Runtime and GPU Usage}
\begin{figure}[h!]
  \centering
  \begin{minipage}[t]{0.40\textwidth}
    \centering
    \includegraphics[width=\linewidth]{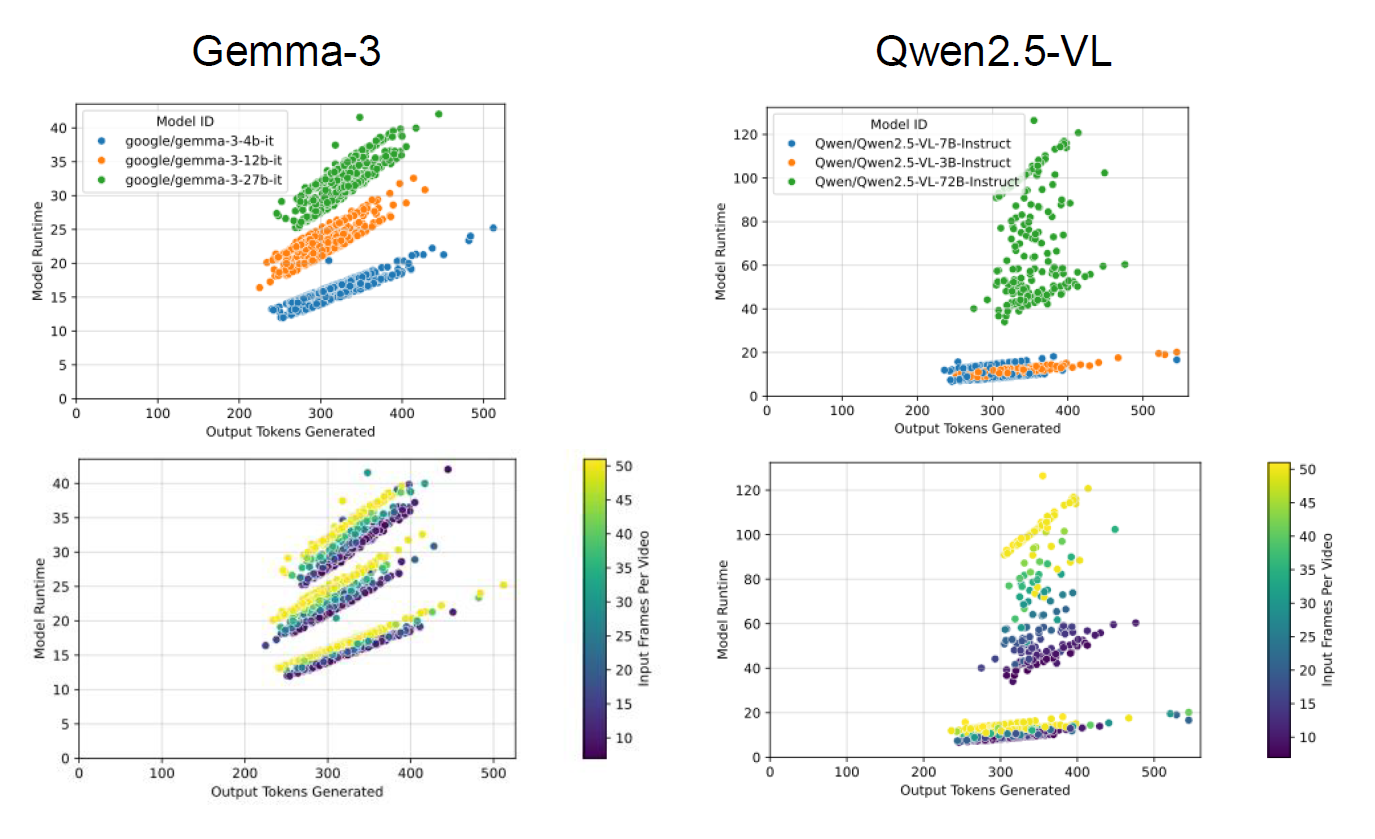}
  \end{minipage}\hfill
  \begin{minipage}[t]{0.30\textwidth}
    \centering
    \includegraphics[width=\linewidth]{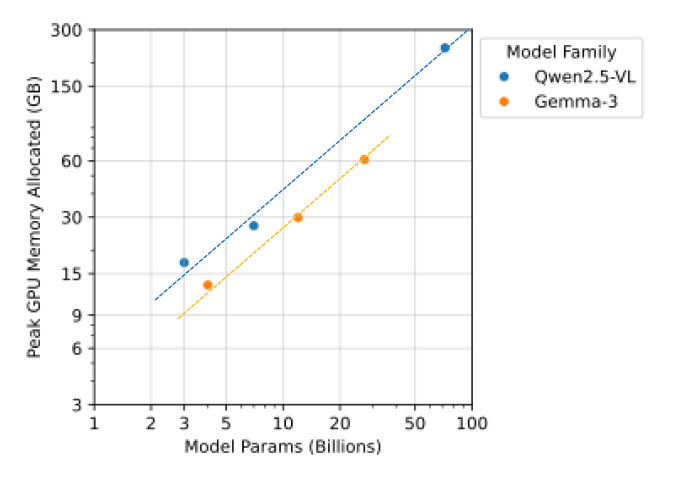}
  \end{minipage}
  \caption{(Top) Correlation of runtime, output tokens generated and input frames per video. (Bottom) GPU memory usage across models. All models fit on a single H100 (96GB) except Qwen2.5-VL 72B, which required 3× H100 GPUs. We use HuggingFace to run our experiments. Total compute is approximately 80 GPU hours.}
  \label{fig:runtime-gpu}
\end{figure}

\noindent We record input frames, runtime statistics such as generated tokens, GPU memory consumption, and other per-model metrics, summarized in \autoref{fig:runtime-gpu}.

\section{Ethical Implications}
\noindent This research used the publicly available PoliTok-DE dataset~\cite{b30}, which contains social media posts from TikTok.
We used this dataset exclusively for model evaluation purposes and did not train any models on this data. 
All data processing was performed locally, and no portions of the dataset are published in our evaluation code, which we make publicly available for reproducibility. 
We acknowledge that content moderation involves inherently subjective judgments influenced by cultural and social context. While our work aims to improve automated content moderation systems by accounting for human label variation, we emphasize that human oversight remains essential in real-world deployment of such systems.

\vspace{12pt}
\color{red}
\end{document}